\documentclass[sn-mathphys]{sn-jnl}% Math and Physical Sciences Reference Style
\jyear{2021}%

\theoremstyle{thmstyleone}%
\theoremstyle{thmstyletwo}%

\usepackage[symbol]{footmisc}
\renewcommand{\thefootnote}{\fnsymbol{footnote}}

\usepackage{lipsum}

\newcommand\blfootnote[1]{%
  \begingroup
  \renewcommand\thefootnote{}\footnote{#1}%
  \addtocounter{footnote}{-1}%
  \endgroup
}

\theoremstyle{thmstylethree}%
\newtheorem{definition}{Definition}%
\newtheorem{problem}{Problem}

\begin{document}

\title[Trajectory Test-Train Overlap in Next-Location Prediction Datasets]{Trajectory Test-Train Overlap in Next-Location Prediction Datasets}


\author*[1,2]{\fnm{Massimiliano} \sur{Luca}}\email{mluca@fbk.eu}

\author[3]{\fnm{Luca} \sur{Pappalardo}}\email{luca.pappalardo@isti.cnr.it}

\author[2]{\fnm{Bruno} \sur{Lepri}}\email{lepri@fbk.eu}

\author[4]{\fnm{Gianni} \sur{Barlacchi$^{+}$\blfootnote{$^+$ work done prior joining Amazon}}}

\affil[1]{\orgname{Free University of Bolzano}, \orgaddress{\street{Piazza Domenicani, 3}, \city{Bolzano}, \postcode{39100}, \state{Italy}}}

\affil[2]{\orgname{Bruno Kessler Foundation}, \orgaddress{\street{Via Sommarive, 19}, \city{Trento}, \postcode{38123}, \state{Italy}}}

\affil[3]{\orgname{ISTI-CNR}, \orgaddress{\street{Via Moruzzi, 1}, \city{Pisa}, \postcode{56127}, \state{Italy}}}

\affil[4]{\orgname{Amazon Alexa AI}, \orgaddress{\city{Berlin}, \state{Germany}}}


\abstract{Next-location prediction, consisting of forecasting a user's location given their historical trajectories, has important implications in several fields, such as urban planning, geo-marketing, and disease spreading. Several predictors have been proposed in the last few years to address it, including last-generation ones based on deep learning. This paper tests the generalization capability of these predictors on public mobility datasets, stratifying the datasets by whether the trajectories in the test set also appear fully or partially in the training set. 
We consistently discover a severe problem of trajectory overlapping in all analyzed datasets, highlighting that predictors memorize trajectories while having limited generalization capacities. 
We thus propose a methodology to rerank the outputs of the next-location predictors based on spatial mobility patterns. With these techniques, we significantly improve the predictors' generalization capability, with a relative improvement on the accuracy up to 96.15\% on the trajectories that cannot be memorized (i.e., low overlap with the training set).}




\maketitle
\section{Introduction}

%{\color{red} STORIA:

%\begin{itemize}
%    \item C'e una incosistenza su come vengono valutati i modelli di NL e viene spesso adottato il metodo di TR 
%    \item C'e' quindi il forte rischio che ci sia un overlap significativo tra training e test 
%    \item Thus, il rischio che i modelli memorizzino traiettorie invece di generalizzare e' elevato 
%    \item Proproniamo un modo per utilizzare leggi di mobilita' well-known per fare un reranking delle predizioni di tutti i modelli NL as multiclass classification per migliorare la capacita' di generalizzare 
%\end{itemize}

%}
%Predicting individuals' future locations is relevant in multiple applications such as monitoring public health  \cite{barlacchi2017you,canzian2015trajectories}, well-being  \cite{pappalardo2016analytical,voukelatou2020measuring}, and traffic congestions \cite{shi2019survey}, and to improve travel recommendation,  geomarketing, and link prediction in social network platforms \cite{zhu2015modeling,burbey2012survey,wu2018location,zheng2018survey,zhao2020event}. 
%Next-Location predictors (NLs) may help policymakers organize the public transportation network. Also, they may be employed by urban planners to decide a city's future developments. Moreover, NLs may help transportation companies provide citizens with a better service in terms of traffic reduction and ease of mobility. 

Next-location prediction is the task of forecasting which location an individual will visit, given their historical trajectories. 
It is crucial in many applications such as travel recommendation, and optimization \cite{shi2019survey,khaidem2020optimizing}, early warning of potential public emergencies \cite{barlacchi2017you,canzian2015trajectories,pappalardo2016analytical,voukelatou2020measuring}, location-aware advertisements and geomarketing, and recommendation of friends in social network platforms \cite{zhu2015modeling,burbey2012survey,wu2018location,zheng2018survey,zhao2020event}. 
Predicting an individual's location is challenging as it requires capturing human mobility patterns \cite{barbosa2018human,luca2020survey} and combining heterogeneous data sources to model multiple factors influencing human displacements (e.g., weather, transportation mode, presence of points of interest and city events). 

The striking development of Deep Learning (DL) and the availability of large-scale mobility data has offered an unprecedented opportunity to design powerful next-location predictors (NLs) and has driven test-set performance on mobility data to new heights \cite{luca2020survey}.
However, little work has been done on how challenging these benchmarks are, what NLs learn, and their actual generalization capabilities. 
Although some studies investigate the predictability of human whereabouts and its relationship with the trajectories' features \cite{song2010limits,amichi2020understanding}, we know comparatively little about how the individuals' trajectories are distributed in mobility benchmarks, making it hard to understand and contextualize our observed results.
Recent studies in natural language processing \cite{lewis2020question,sen2020models} and computer vision \cite{zhou2021domain} show that DL models excel on specific test sets but are not solving the underlying task. 
In this paper, we investigate whether it is the case for NLs too.

We perform an extensive study of the test sets of several public next-location benchmark datasets \cite{luca2020survey} and evaluate a set of state-of-the-art DL-based NLs on their generalization capability.
We identify three levels of generalization that an NL should exhibit: (i) \emph{known mobility}, requiring no generalization beyond recognizing trajectories seen during the training phase; (ii) \emph{fragmentary mobility}, requiring generalization to novel compositions of previously observed trajectories; and (iii) \emph{novel mobility}, requiring generalization to a sequence of movements not present in the training set.
It is unclear how well state-of-the-art NLs perform on each of these three scenarios. 

To address this compelling issue, we stratify mobility data by whether the trajectories in the test set also appear fully or partially in the training set. 
We quantify the overlap between trajectories with three measures accounting for different ways of computing the percentage of locations in the test trajectories that are also in the training trajectories. 

We find that, in five next-location benchmark datasets, there is a severe problem of trajectory overlapping between the test and training sets when composing them randomly: $\sim$ 43\% to 72\% of test trajectories overlap at least with 50\% of the points with trajectories in the training set, and with 7\% to 14\% of test sub-trajectories entirely overlap training sub-trajectories. 
In other words, based on the standard way training and test sets are split in the literature, a significant portion of the trajectories in the test sets have already been seen during training. 
%These results imply that 21\% to 46\% of the test set reveals how well the model predicts the next location with unseen trajectories. The other only probes for how well NLs can memorize trajectories seen during training. 

Based on these observations, we propose to evaluate NLs on \emph{stratified test sets based on the overlap between trajectories in the training set}. 
We find significant variability in model performance, varying the percentage of overlap.
Indeed, we find an accuracy $\leq 5\%$ when predicting unseen trajectories (novel mobility) and $\geq 90\%$ when predicting trajectories with high overlaps (known mobility). 
Surprisingly, we also find that DL-based NLs perform even worse than baseline models (e.g., Mobility Markov Chain or MMC \cite{gambs2012next}) when tested on novel mobility.
Our results are consistent across the datasets analyzed and the NLs selected, demonstrating that current train/test splits are flawed, and more robust methods are needed to evaluate the generalization capabilities of NLs.  
We also show a way to improve next-location prediction accuracy, especially for the novel mobility scenario, injecting mobility laws into state-of-the-art NLs through a learning-to-rank task.
In a nutshell, this paper provides the following novel contributions:
\begin{itemize}
    \item We show that standard train/test splits of trajectory datasets generate a high trajectory overlap, proposing three metrics to quantify it;
    
    \item We evaluate NLs on stratified test sets and show that DL-based NLs do not generalize well on novel mobility, being outperformed by other simpler baselines (e.g., Mobility Markov Chains);
    
    \item We show how to improve the accuracy of DL-based NLs, especially for the novel mobility behavior, by performing a rerank of the models' scores based on spatial mobility patterns; 

\item Based on our findings, we provide a list of recommendations to improve datasets' creation and models' evaluation for next-location prediction.
    
%    \item We made available the code to measure trajectories' overlap as well as to stratify test data.
\end{itemize}

\section{Related Work}

\subsection*{Model Generalization} 
Measuring the generalization capabilities of deep neural networks has recently captured the attention of researchers in artificial intelligence. 
Lewis et al. \cite{lewis2020question} find that, in popular Question Answering (QA) datasets, 30\% of test-set questions have a near-duplicate in the training sets and that all models perform worse on questions that cannot be memorized from training sets. Sen and Saffari \cite{sen2020models} show that QA models do not generalize well on unseen question-context pairs. However, they still perform well on popular QA benchmarks because of their high overlap between train and test data. Liu et al. \cite{liu2021challenges} go beyond the data and study the key factors that impact generalization in QA. 
An essential impact in generalization is played by cascading errors from retrieval, question pattern frequency, and entity frequency.

\subsection*{Predictability of Human Mobility}
%In the context of human mobility, this is the first time, to the best of our knowledge, that the generalization capabilities of deep learning models are discussed. 
Several studies measure the limits of predictability of human mobility \cite{barbosa2018human, luca2020survey}. 
%In general, some types of trajectories are more predictable than others and can be easily memorized while for others, we need models able to generalize. 
Song et al. \cite{song2010limits} analyze mobility traces of anonymized mobile phone users to find that 93\% of the movements are potentially predictable. 
Zhang et al. \cite{zhang2022beyond} show that, when considering the mobility context (e.g., visiting time, kind of place visited), the upper bound of potential predictability in human mobility increases. 
Other studies show that this upper bound depends on the data scale and the processing techniques adopted \cite{smolak2021impact, kulkarni2019examining,hofman2017prediction}.
%Other studies focus on the behavior of the individuals and its role on predictability \cite{amichi2020understanding,do2021estimating}. 
In \cite{amichi2020understanding,do2021estimating}, there are shreds of evidence that the so-called explorers (e.g., individuals without a routinary behavior) \cite{pappalardo2015returners} are less predictable than the others. 
All the works discussed suggest that models may memorize certain trajectories (e.g., routinary mobility) while not being able to generalize well on novel mobility (i.e., mobility not observed during the training phase).

\subsection*{Next-Location Prediction} 
%A seminal work proposes a probabilistic model combining people's trajectories and geographical features such as land use, POIs, and distance of trips \cite{calabrese2010human}. Another approach clusters GPS data into meaningful locations and incorporate them into a Markov model to predict individuals' future movements \cite{ashbrook2002learning}. 
Most NLs are based on (gated) recurrent neural networks (RNNs). RNNs \cite{hinton1986rnn} can efficiently deal with sequential data such as time series, in which values are ordered by time, or sentences in natural language, in which the order of the words is crucial to shaping its meaning. 
In Spatial Temporal Recurrent Neural Networks (ST-RNN) \cite{liu2016predicting}, RNNs are augmented with time- and space-specific transition matrices.
Through linear interpolation, each RNN layer learns an upper and lower bound for the temporal and spatial matrices, which are then used to infer an individual's next visited location.
Long Short-Term Memory Projection (LSTPM) \cite{sun2020go} use sequential models to capture long- and short-term patterns in mobility data. 
The authors rely on a non-local network \cite{wang2018non} for modeling long-term preferences and on geo-dilated RNNs inspired to capture short-term preferences \cite{chang2017dilated}.
More sophisticated models like DeepMove \cite{feng2018deepmove} use attention layers to capture the periodicity in mobility data. 
First, past and current trajectories are sent to a multi-modal embedding module to construct a dense representation of spatio-temporal and individual-specific information. Next, an attention mechanism extracts mobility patterns from historical trajectories, while a Gated Recurrent Unit (GRU) handles current trajectories. Finally, the multi-modal embedding, GRU, and attention layer outputs are concatenated to predict the future location.
Recently, Spatio-Temporal Attention Network (STAN) \cite{luo2021stan} proposes to capture spatio-temporal information to leverage spatial dependencies explicitly. In particular, the authors use a multi-modal embedding layer to model historical trajectories and the GPS locations in the current trajectories. The embeddings are then forwarded to a spatio-temporal attention mechanism that selects a set of potential next locations.
Many other works deal with spatio-temporal data using (gated) RNNs and attention mechanisms. Some of them also deal with the semantic meaning associated with locations. Examples of such models are Semantics-Enriched Recurrent Model (SERM) \cite{yao2017serm}, Hierarchical Spatial-Temporal Long-Short Term Memory (HST-LSTM) \cite{kong2018hst}, VANext \cite{gao2019predicting}, and Flashback \cite{yang2020location}.
Other Deep Learning solutions to next-location prediction have been discussed in a recent survey paper \cite{luca2020survey}.
%Next-location prediction has been widely explored prior to the deep learning explosion using probabilistic or pattern-based approaches, which can work with a reasonably small amount of data \cite{burbey2012survey,zheng2018survey}. 

%Gambs et al. \cite{gambs2010show} introduce a Mobility Markov Chain (MMC) in which states represent POIs and transitions between states correspond to a movement between two POIs \cite{gambs2010show,gambs2012next}.

%Later on, there has been an increasing number of works that address next-location prediction and deal with spatio-temporal data. 

\section{Problem Definition}
Next-location prediction is commonly defined as the problem of predicting the next location an individual will visit given their historical movements, typically represented as spatio-temporal trajectories \cite{luca2020survey}. 
%Formally, we define a trajectory as follows. 

\begin{definition}[Trajectory]
A spatio-temporal point $p=(t, l)$ is a tuple where $t$ indicates a timestamp and $l$ a geographic location. 
A trajectory $P = p_1,p_2,\dots,p_n$ is a time-ordered sequence of $n$ spatio-temporal points visited by an individual, who may have several trajectories, $P_{1}, \dots, P_{k}$, where all the locations in $P_i^u$ are visited before locations in $P_{i+1}$.
\end{definition}

Given this definition, we formalize next-location prediction as follows:

\begin{problem}[Next-location prediction]
\label{problem_def}
Given the current trajectory of an individual $P_{k} = p_1,p_2,\dots,p_n$ and their historical trajectories $\mathcal{H} = P_{1}, \dots, P_{k-1}$, next-location prediction is the problem of forecasting the next point $p_{n+1} \in P_{k}$. 
\end{problem}

In other terms, a next-location predictor (NL) is a function $\mathcal{M}(P_k, \mathcal{H}) \rightarrow p_{n+1}$, which takes the current trajectory $P_k$, the set of $u$'s historical trajectories $\mathcal{H}$, and returns a spatio-temporal point $p_{n+1}$ in $P_k$.

\section{Trajectory Overlap}
%We refer to trajectory overlap as a way to measure the percentage of test data already seen in the training phase. 
An NL should be able to predict an individual's next location in three scenarios: \emph{(i)} the NL has seen the individual's entire current trajectory during the training phase; \emph{(ii)} it has seen the current trajectory only partially, or it has seen a very similar trajectory of the same individual; \emph{(iii)} the current trajectory was absent from the training set.
The latter scenario is essential, as machine learning models' ability to generalize is their capacity of making predictions on data never seen during the training phase \cite{kawaguchi2017generalization}.

%\begin{itemize}
%    \item NLP dovrebbe prevedere sia routine che out of routine
%    \item però nella mobilità spesso c'è overlap
%    \item quindi può essere che non siamo bravi nell'out of routine
%    \item spesso però, nella valutazione (in particolare in quella trajectory reconstruction) l'out of routine non è proprio valutata per come è definita la valutazione. come sono compoasti train e test.
%    \item la valutazione è molto biased dalla presenza di subtrajectories
%    \item sarebbe auspicabile valutare i modelli a diversi soglie di overlap, per verificare la loro capacità di generalizzare e quindi di prevedere la mobilità out of routine.
%\end{itemize}
However, in next-location prediction, there may be a significant \emph{overlap} between trajectories in the test set and those in the training set. 
For example, some test and training trajectories may belong to the same individual. 
Since human mobility is routinary, an individual's trajectories are similar to each other \cite{barbosa2018human,schlapfer2021universal}, leading to scenarios \emph{(i)} and \emph{(ii)} above.
Given this discussion, we investigate \emph{the extent to which the overlap between trajectories in the test and training sets influences the model's ability to generalize}.
We explore three ways to examine overlap: Jaccard Similarity (JS), Longest Common Subsequence (LCST), and Overlap From the End (OFE).
%, and each of them captures different types of overlap.

%In particular, given a trajectory $P \in D_{\text{test}} = p_1, p_2, \dots p_n$, we are interested in know

Jaccard Similarity (JS) measures the percentage of locations in the test trajectories that are also in the training trajectories, regardless of the order in which locations appear. 
Test trajectories with a high JS have many locations in common with training trajectories. 
In contrast, test trajectories with low JS should be less predictable as they are mainly composed of locations that are not in the training trajectories. 
Formally, we define the JS between a trajectory $R \in D_{\text{test}}$ and $P \in D_{\text{train}}$ as: 
$$\text{JS}(R, P) = \frac{\lvert P \cup R \rvert -\lvert P\cap R \rvert}{\lvert P \cup R \rvert}$$
We quantify the overlap between $R$ and the training set as the maximum JS over all the trajectories in the training set:
$$ \max_{P \in D_{\text{train}}} \text{JS}(R, P).$$

JS $\in [0, 1]$, where 1 indicates a full overlap (all locations in $R$ are at least in a trajectory in $D_{\text{train}}$) and 0 indicates no overlap (none of the locations in $R$ are in the training set).

%A limitation of the JSC is that the measured overlap does not consider the temporal and spatial order in which the locations are visited. Moreover, it does not handle more than one visit to the same place.
%To overtake these limits, we also define the following overlap measures.

The Longest Common SubTrajectory (LCST) is the longest subtrajectory in common between two trajectories. 
%In our case, given a trajectory in the test set, we aim to find the longest common subsequence matching in any training trajectory. 
Formally, given a training trajectory $P = p_1,p_2,\dots,p_n $ and a test trajectory $R= r_1,r_2,\dots,r_m$, we define a recursive function $f(P, R)$ as: %for each possible $P \in D_{\text{train}}, R \in D_{\text{test}}$ as 
$$\small
    f(P, R) = \begin{cases} 
        0, & \mbox{if } i=0 \mbox{ or } j=0 \\ 
        f(p_{i-1}, r_{j-1})+1, & \mbox{if } i,j {>} 0 \mbox{ and } p_{i} {=} r_{j} \\
        \mbox{max}(f(p_{i-1}, r_{j}), f(p_{i}, r_{j-1})) & \mbox{if } i,j {>} 0 \mbox{ and } p_{i} {\neq} r_{j}
    \end{cases}
$$

\noindent where $|P|$ and $|R|$ indicate the length of the training and test trajectories, respectively, and $f(P, R) \in [0, \min(|P|, |R|)]$. 
The LCST between $P$ and $R$ is then: 
$$\text{LCST}(P, R) = f(P, R) /\ |R|.$$

We quantify the overlap between R and the training set as the maximum LCST over all the trajectories in the training set: 
$$ \max_{P \in D_{\text{train}}} \text{LCST}(R, P).$$

%LCST takes into account the spatial and temporal order in which visits take place. 
%However, we are looking for a common sequence regardless the position in the trajectory and it may be the case that the last visited places (e.g., last elements of the trajectories) are more relevant than the previous ones.

The Overlap From End (OFE) enforces that the common subtrajectory is at the end of the two trajectories. 
Formally, given a trajectory $P = p_1, p_2, \dots, p_n$, we define $P' = p_n, \dots, p_2, p_1$ as its reversed trajectory.
We then compute $\text{OFE}(R, P)$ with Algorithm \ref{alg:ofe} and
quantify the overlap between $R$ and the training set as the maximum OFE over all the trajectories in the training set: 
$$ \max_{P \in D_{\text{train}}} \text{OFE}(R, P).$$
In other terms, given a trajectory in the test set, we scan all the trajectories in the training set and we compute, for each pair $(P,R)$, we compute, starting from the last point the number of common points. We than convert this number into a percentage. Finally, the OFE of $P$ is the higher percentage found. 
\begin{algorithm}
\caption{OFE Computation}\label{alg:ofe}
\begin{algorithmic}
\State $\text{overlaps} \gets \text{dictionary()}$
\For{$R' \in D_{\text{test}}$}
\State $\text{overlap} \gets 0$
\For{$P' \in D_{\text{train}}$}
\State $\text{count} \gets 0$
\For{$k \in \{ 0, \dots, \text{min}(|P'|, |R'|)\}$}
\If{$R'[k] = P'[k]$}
\State $\text{count} \gets \text{count}+1$
\ElsIf{$R'[k] \neq P'[k]$}
\textbf{Break}
\EndIf 
\EndFor
\If{$\text{count}  /\  |R'| > \text{overlap}$}
\State $\text{overlap} \gets \text{count}  /\  |R'|$
\EndIf
\EndFor 
\State $\text{overlaps}[R'] \gets \text{overlap}$
\EndFor
\end{algorithmic}
\end{algorithm}

%\subsection{Recurrent Neural Networks}

\section{Experimental Setup}

\subsection{Datasets}
We use five public datasets widely adopted in the literature to evaluate NLs \cite{luca2020survey} (see Table \ref{tab:data_summary}).
Three of them (Gowalla, Foursquare New York, Foursquare Tokyo) are collected through social networking platforms, in which mobility traces are generated by the users' georeferenced posts (check-ins). 
Consequently, these mobility traces are sparse both in time and space. 
The other two datasets (Taxi Porto and Taxi San Francisco) describe GPS traces from taxis dense in space and time.
In detail, Gowalla was a location-based social network platform that, like Foursquare, allowed users to check-in at so-called spots (venues) via a website or an app. The dataset \cite{cho2011friendship} has almost six million check-ins collected over a year and a half, from February 2009 to October 2010. Each check-in contains the user identifier, location identifier, latitude and longitude pair, and timestamp. The dataset also contains information on the users' friendship network, which has around 200,000 nodes and one million edges.
Foursquare is another location-based social network platform that allows users to check in into places. Data can be collected through the available APIs. A widely used dataset based on Foursquare is described in \cite{yang2014modeling}. The information contained are the same as Gowalla, with additional information about the category of the venue.
Piorkowski et al. \cite{taxi_sf} collected taxi trajectories in San Francisco in May 2008. Each point in a trajectory includes the taxi's identity, latitude, longitude, timestamp, and occupancy. Points are sampled every 10 seconds on average.
Moreira et al. \cite{moreira2013predicting} (ECML/PKDD Challenge) collected taxi trajectories in Porto, Portugal. For each trajectory, we have the taxi's identifier, the latitude, longitude, and timestamp showing when the trip began. For each trajectory, data are sampled every 15 seconds.
The dataset also includes auxiliary information for each trip, such as the trip's typology (e.g., sent from the central, demanded to the operator, demanded to the driver), the stand from which the taxi left, and a phone number identification for the passenger.

To extract trajectories from these datasets, we follow the same approach as in \cite{feng2018deepmove}: first, we filter out the users with less than ten records; second, we cut the sequence of records into several trajectories for each user based on the time interval between two neighbor records. 
As in \cite{feng2018deepmove}, we choose 72 hours as the default interval threshold based on the practice. Finally, we remove the users with less than five trajectories.

%In the subsequent sections, we present results for Foursquare NYC and Taxi Porto only, leaving the results for all other datasets (Gowalla, Foursquare Tokyo, Taxi SF) to the Appendix.

\subsection{Models}
%Recurrent Neural Networks (RNNs) \cite{hinton1986rnn} are the building blocks of the vast majority of NLs \cite{luca2020survey} because they can efficiently deal with sequential data and spatio-temporal patterns. %For example, ST-RNN \cite{liu2016predicting} and DeepMove \cite{feng2018deepmove}, two seminal works in this field, are based on recurrent neural networks. 

%An RNN consists of a sequence of gates, each producing a hidden state based on the current input and the output from the previous gate. A limitation of RNNs is the so-called vanishing gradient problem \cite{kolen2001gradient}: they may have difficulties in propagating information found at early steps, losing relevant information at the beginning of a sequence when it is time to analyze its end \cite{kolen2001gradient}. Long-Short-Term Memory networks (LSTMs) \cite{hochreiter1997long} and Gated Recurrent Units (GRUs) \cite{cho-etal-2014-learning} are two gate implementations that mitigate this problem. 

%Since RNNs, LSTMs, and GRUs are the building blocks of most NLs \cite{luca2020survey}, we focus on examining their generalization capabilities, assuming that if they fail to generalize, so do more sophisticated models based on them.

We validate our hypothesis by testing the generalization capability of the following state-of-the-art DL-based NLs.
\begin{itemize}
    \item \textbf{RNN} \cite{hinton1986rnn}, the building block of the majority of NLs. RNNs are commonly adopted to model sequential data such as time series and natural language, in which the order of the items is crucial to shaping its meaning. RNNs are also widely used as building blocks of NLs to capture spatial and temporal patterns in the trajectories. 
    An RNN is made of a sequence of gates, each one outputting a hidden state $h_i$ based on the current input $x_i$ and the previous gate $h_{i-1}$. 
    In this work, a gate is implemented as a hyperbolic tangent function (tanh).
    \item \textbf{ST-RNN} \cite{liu2016predicting} enhances RNNs with time- and space-specific transition matrices in this study.
    Each RNN layer learns an upper and lower bound for the temporal and spatial matrices via linear interpolation. These matrices are then used to predict where a person will go next.
    \item \textbf{Deep Move}  \cite{feng2018deepmove} uses attention mechanisms to capture spatio-temporal periodicity in the historical trajectories. 
    Also, the model uses GRUs (gated RNNs) to capture patterns in the current trajectory and relies on a multi-modal embedding to capture individual preferences and project trajectories in a low-dimensional space before passing them to the attention mechanisms and GRUs.
    \item \textbf{LSTPM} \cite{sun2020go} combines long- and short-term sequential models: long-term patterns are modeled using a non-local network   \cite{wang2018non}, short term preferences are captured using a geographic-augmented version of the concept of dilated RNNs \cite{chang2017dilated}.
    \item \textbf{STAN} explicitly captures spatio-temporal information using a multi-modal embedding to represent the trajectories and a spatio-temporal attention mechanism to capture patterns in the data \cite{luo2021stan}. The role of the attention mechanisms, supported by a balanced sampler, is to rank potential next locations. 
\end{itemize}

\subsection{Training}
We split the trajectories into a training set, a validation set, and a test set for each dataset. 
All sets include trajectories from several users.
We sort the trajectories temporally for each user and put the first 70\% in the training set, the following 10\% in the validation set, and the remaining 20\% in the test set. 
%The test set is stratified according to the percentage of overlap with the training set as described in the subsequent sections.

All models are implemented with PyTorch and are made available through the library LibCity \cite{libcity}. 
We follow the same configuration as \cite{feng2018deepmove} and use Adam \cite{kingma2014adam} as optimizer.

We ran the experiments on a machine with 126GB of memory and two Nvidia RTX 2080Ti.

%\begin{table}[]
%\centering
%\resizebox{\linewidth}{!}{%
%\begin{tabular}{l|l|l|l}
%\textbf{Hyperparameter} & \textbf{Value} & \textbf{Hyperparameter} & \textbf{Value} \\ \hline
%Dropout probability     & 0.3            & L2                      & 1 * 1e-5       \\
%LR                      & 5 * 1e-4       & Gradient Clipping       & 5.0            \\
%LR step                 & 2              & LR decay                & 0.1           
%\end{tabular}%
%}
%\caption{Default hyperparameters values of our models. 
%We fine-tune them on each dataset and obtain almost the same values.
%}
%\label{tab:hyper}
%\end{table}

\begin{table}[]
\centering
\small
%\resizebox{\textwidth}{!}{%
\begin{tabular}{ll|llll}
            &                              &  Users & Locations & Trajectories \\ \hline
Gowalla     &  \cite{cho2011friendship}      & 5300  & 125,771   & 72,593                 \\
Foursquare NYC &  \cite{yang2014modeling}       & 4390  & 13,960    & 12,519                 \\
Foursquare Tokyo    & \cite{yang2014modeling}        & 935   & 21,394    & 34,662                 \\
%FS DeepMove & \cite{feng2018deepmove}       & 2108  & 10,497    & 10,497                 \\
Taxi Porto  & \cite{moreira2013predicting}   & 500   & 8524      & 94,214                 \\
Taxi SF     & \cite{taxi_sf}              & 500   & 9321      & 103,120                \\ \hline
\end{tabular}%
%}
\caption{Properties of the datasets adopted in our study. We describe each dataset's time span, number of users, number of locations, and the number of trajectories extracted.}
\label{tab:data_summary}
\end{table}

\section{Testing Generalization Capability}

We evaluate the performance of all models using the $k$-accuracy (ACC@$k$), the most common evaluation metric in the literature \cite{luca2020survey}. 
NLs output a list of all possible locations an individual will visit next ranked from the most to the least likely. 
ACC@$k$ indicates how many times the true location is among the $k$ top predicted locations.
We evaluate all models using ACC@5. 

We compare the DL models with Mobility Markov Chains (MMCs) \cite{gambs2012next}, in which the visited locations are the states of a Markov chain and a transition matrix represents the first-order transition probabilities between these locations. 
The choice of MMCs as a baseline is justified because they cannot generalize as they summarize the training data. 

\begin{figure*}[htb!]
    \centering
    \includegraphics[width=1\linewidth]{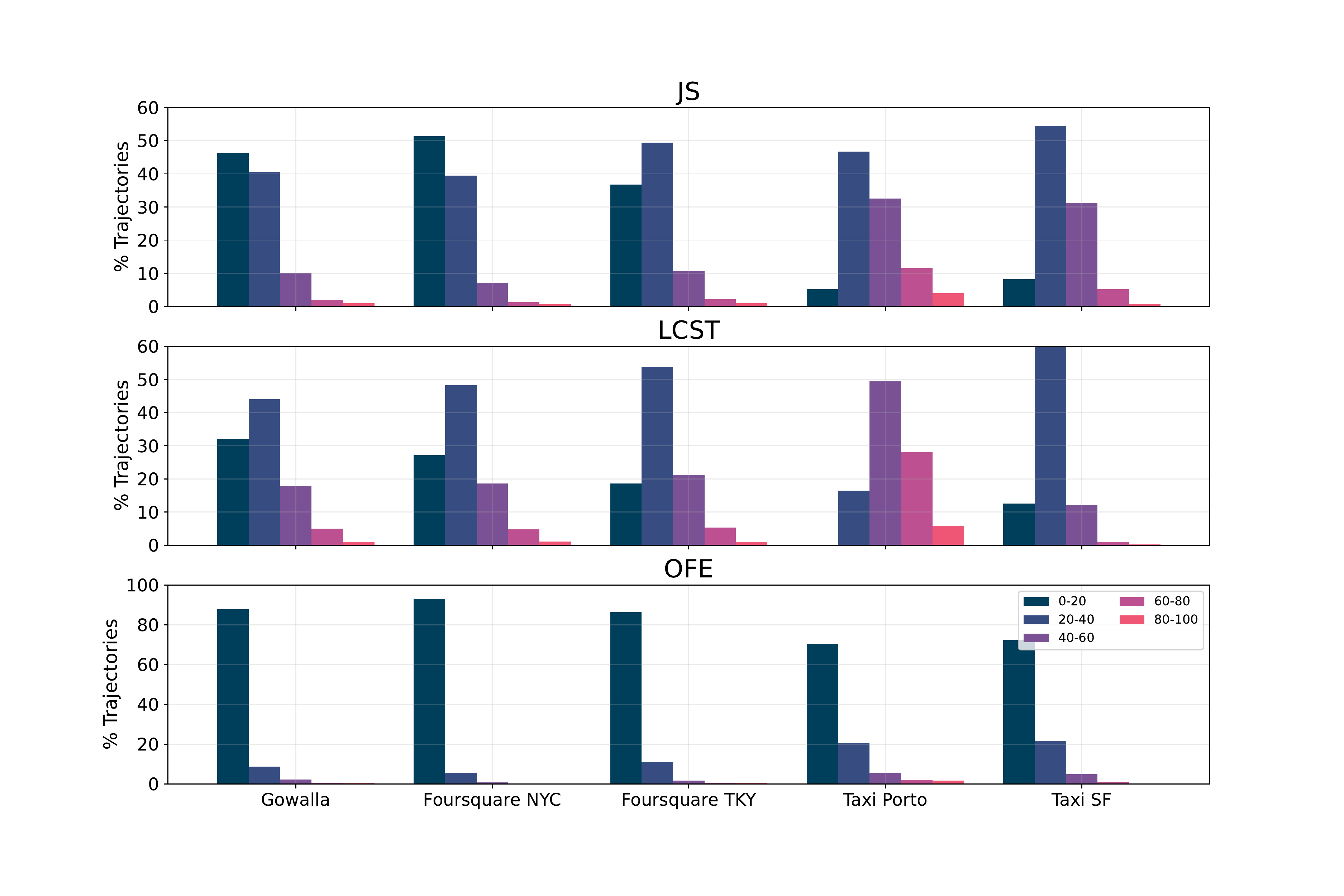}
    \caption{Fraction of the test trajectories with an overlap of 0-20\%, 20.40\%, 40-60\%, 60-80\%, and 80-100\% with the training trajectories, for the all datasets, for the evaluation metrics JS, LCST, and OFE.}
    \label{fig:overlap_stats}
\end{figure*}

%\subsection{Trajectory Overlap}
For all datasets and overlap metrics (JS, LCST, and OFE), we compute the number of trajectories in the test set with an overlap with the training set between 0-20\%, 20-40\%, 40-60\%, 60-80\%, and 80-100\%. 
Figure \ref{fig:overlap_stats} shows the results for all the datasets analyzed. 

The percentage of trajectories with a high overlap (between 80\% and 100\%) varies widely with the overlap metric and the dataset. 
Taxi datasets have more trajectories with a high overlap than the datasets based on check-ins, suggesting that the overlap problem is more severe in GPS traces than in check-ins. 
We also observe that JS and LCST produce similar overlaps, while with OFE, the number of trajectories with low overlap is remarkably higher. 
This is due to the severe constraints that OFE imposes by definition (e.g., the overlap is evaluated starting only from the end of the trajectory).
%Results in Figure \ref{fig:overlap_stats}. 
%, with the exception of the number of trajectories with an overlap between 40\% and 60\% in Taxi Porto that in PW are significantly higher than in TR. 
%for TR, we have a percentage of trajectories with low overlap than in PW. 
%We face the opposite behavior if we measure the overlap with OFE. 
%Specifically, with OFE, most of the trajectories have a low overlap (within 0-20\%), reaching almost 100\% for PW. 

In any case, Figure \ref{fig:overlap_stats} highlights that a significant overlap exists between the test and the training set, introducing a bias when evaluating NLs using a random train-test split. 
Hence, we investigate to what extent this overlap affects model performance. 

\begin{figure*}[htb]
    \centering
    \includegraphics[width=\linewidth]{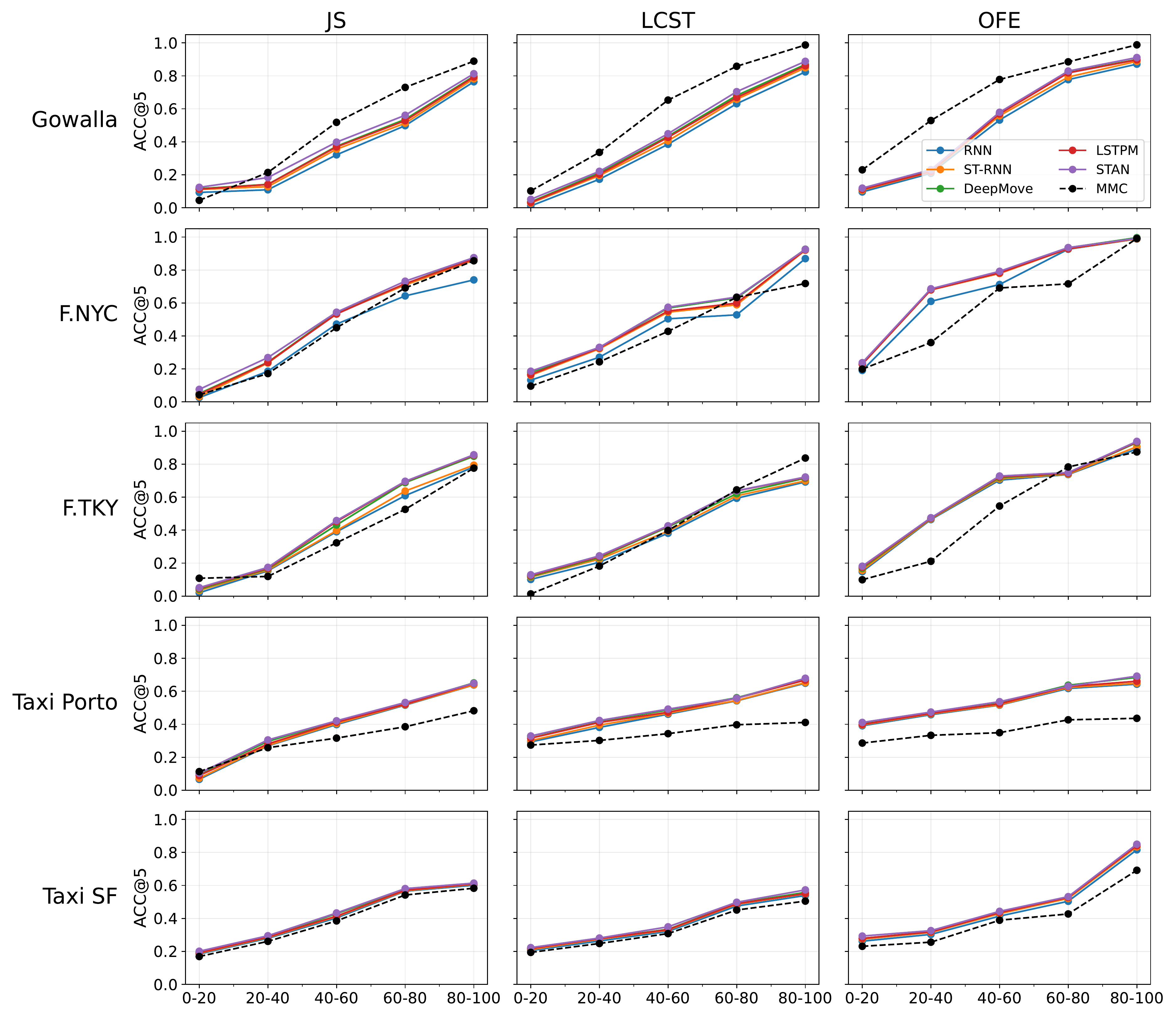}

    \caption{Results (in terms of ACC@5) for all the datasets and models. We compute the accuracy for the three overlap metrics (JS, LCST, OFE)  and for five bins of percentage of trajectory overlap (from 0-20\% to 80-100\%).}
    \label{fig:performances}
    
\end{figure*}

Figure \ref{fig:performances} shows the performances for all the NLs and overlap metrics. 
Here, increasing the overlap induces a striking improvement in the model performance for both MMC and the NLs, which have similar performance.
We present all performances in detail in the Supplementary \ref{app:perf}.
%The latter can be justified as, by definition, MMCs are memory-less models and, therefore, there should not be a significant difference when predicting all the points in a trajectory and only the last one. 

%Regarding DL models, in this plot it is also clear that performances are biased by the overlap problem. 
For example, for Foursquare New York and OFE, the performance of NLs is close to 100\% on a test made of trajectories with an overlap with the training set in the range 80-100\%. 
Results for Taxi Porto follow a similar increasing trend, although with less striking performance.

Overall, Figure \ref{fig:performances} shows that model performance is strongly affected by trajectory test-train overlap, suggesting that NLs memorize trajectories instead of generalizing. 
NLs perform well on trajectories with high overlap with the training set but poorly on trajectories with low overlap.
These results raise the question of how to improve the accuracy of NLs for low overlap scenarios.

\section{Learning to Rank Locations Using Mobility Laws}
%Our results highlight the inability of NLs to generalize: models perform well on trajectories with high overlap with the training set but poorly on trajectories with low overlap. 
%We propose a way to improve the accuracy on the latter by \emph{reranking} the outputs of uncertain predictions.
%The reranking procedure aims to boost the accuracy on the test trajectories with low overlap with the training ones. 
%Indeed, the main goal is to provide a methodology to improve the generalization capabilities of NLs.

A possible reason why NLs perform poorly on trajectories with low overlaps lies in the type of DL tools they rely on, i.e., RNNs: they focus on memorizing regularities in long sequences, thus limiting NLs' generalization capabilities. 
Wrong location predictions happen when the probabilities assigned to each potential location by the NL (i.e., the locations' scores) are low and relatively uniformly distributed. 
Our intuition is to rerank locations based on new scores obtained, injecting human mobility laws into NLs.
We select three prominent human mobility laws \cite{barbosa2018human,luca2020survey}: 
\begin{itemize} 
\item the distance law \cite{barbosa2018human}: people prefer travelling short distances. 
Given an individual's trajectories $P=p_1,p_2,\dots,p_n$, we compute the Haversine distance between all the consecutive locations $p_i, p_{i+1}$ and consider the average of the distances as a feature $dist_u$;   
\item the visitation law \cite{schlapfer2021universal}: the visits to a location decrease as the inverse square of the product of their visiting frequency and travel distance. 
We denote as $f$ the number of visits to a location (by any individual) and compute how many people visit it within a distance $r$. 
An individual's probability to visit location $p_{i+1}$ is given by a power-law of the form $p_{i+1}(r,f) = \mu_i/(rf)^\gamma$, with $\gamma=1.6$, a parameter fitted with the least squares method. 
We use the five most probable locations $top_n, n \in 1 \dots 5$ as an input to the reranker.
\item the returner and explorer dichotomy \cite{pappalardo2015returners}: individuals naturally split into two profiles based on their degree of spatial exploration.  
We compute the average radius of gyration $r_g(u)$ and the 2-radius of gyration $r_{g}^{(2)}(u)$ for each individual and compute the ratio $\frac{r_g(u)}{r_{2g}(u)}$ using the scikit-mobility \cite{pappalardo2019scikit} library. The profile of the user is then translated into a binary feature: 0 if the individual is a returner and 1 if the individual is an explorer. We denote this feature as $re_u$. \end{itemize} 
%We then re-rank the locations predicted by an NL using the following three-steps procedure: 
%\\ \noindent %\textbf{NL training:} we train an NL  obtaining an array $A$ of probabilities (scores). 
%The size of $A$ is the number of all possible locations.
{\color{black}}
%\\ \noindent \textbf{Reranker training:} 
%We use a fully connected neural network to train a binary classifier to rerank the locations predicted by the model. 
Our approach consists in predicting the next location using a NL, and then combining into a single scoring model, i.e., a fully connected neural network, both the NL score for the location and the mobility laws. We trained the network using the binary cross-entropy loss $\mathcal{L} = - \sum_{i\in\{0,1\}}y_i \log{p(y_i)}$, where $y_i$ is the label (i.e., 0 or 1) and $p(y_i)$ is the predicted probability.
 
%$.
The training dataset consists of vectors of the form $ [ \text{NL}_i(P), dist_u, top_1,\dots, top_5, re_u]$. We denote with $\text{NL}_i(P)$ the score of the NL for a given location $i$ starting from a trajectory $P$. The label is 1 if the location $i$ is the individual's next-location and 0 otherwise. This means that in a dataset with $n$ locations, for each trajectory we have a positive sample (e.g., the correct next location) and $n-1$ negative samples for each trajectory. 
As the number of incorrect samples is much higher than the correct ones, for each correct location, we randomly sampled $k=20$ wrong locations (e.g., locations that are different from the actual next location the individual will visit), as we found it to be a good trade-off between performance and dataset size. %We used 20\% of the training set to optimize the ratio positive:negative.
%for each training trajectory we provide 21 samples: 1 positive sample (the score and other features associated to the correct next location) and information about other 20 incorrect locations that are randomly selected. The FC layer is followed by a softmax that allows us to have the new reweighted score for each location as output of the model.
%The loss function of RR is the binary cross-entropy:
%$$ H_p(q) = -\frac{1}{N} \sum_{i=1}^N y_i log(p(y_i)) + (1-y_i) log(1-p(y_i)) $$
 
%At inference time, we score each location individually, and the new score simply sorts the final ranking. %We compute the updated score for each possible location $i$, and finally, we compute the accuracy using the newly obtained reranked array.
%\\ \noindent \textbf{NL and reranker testing:}
%we test an NL on the test set. We give the same input we use for training to the RR: $\text{RR}(\text{NL}_i(P), dist_u, top\_5, re_u)$ and, as output, we have the newly updated score for location $i$ in a trajectory $P$. 

%\section{Results - Reranking}

\begin{figure*}[htb]
    \centering
    \includegraphics[width=\linewidth]{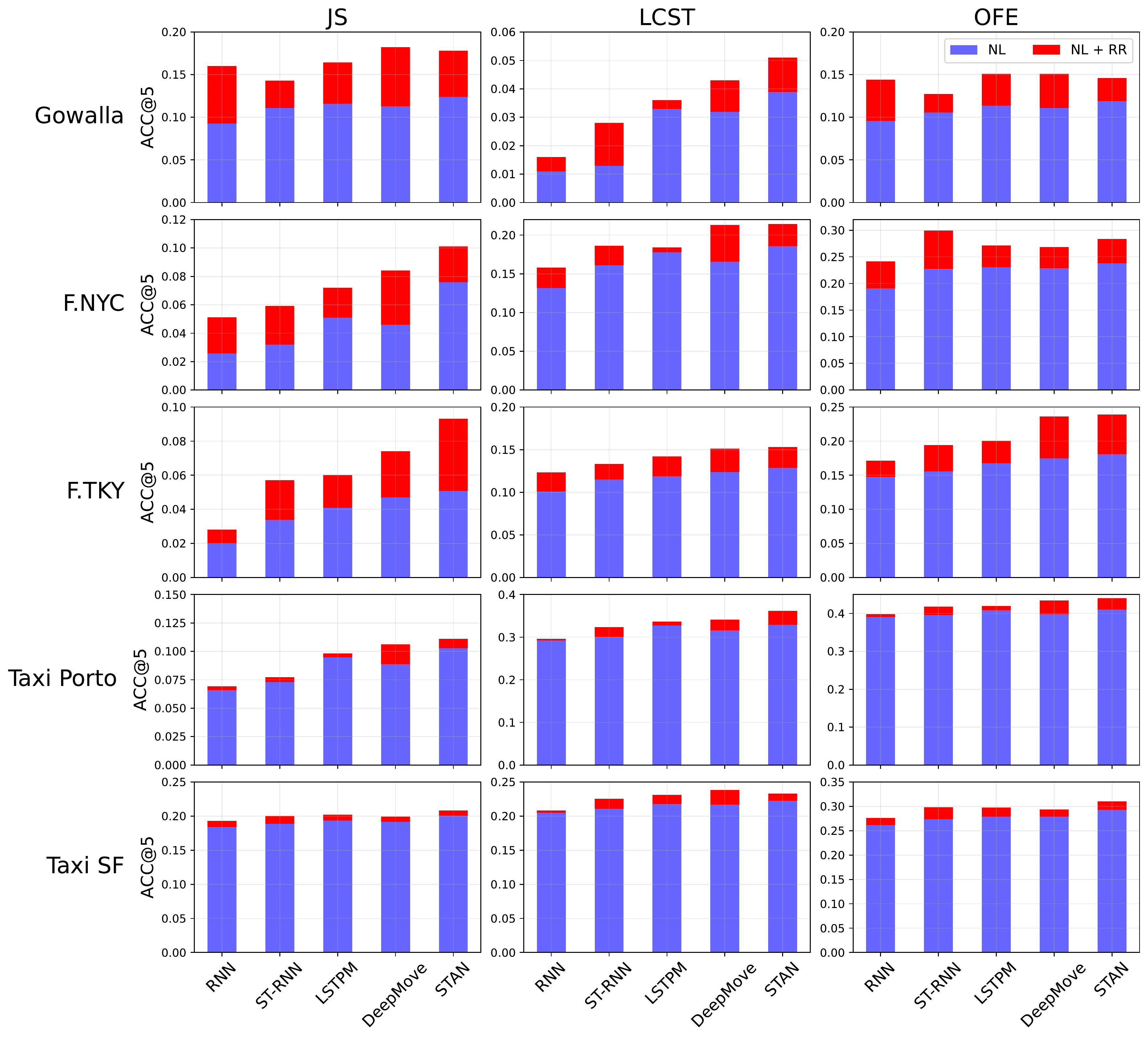}

    \caption{Results (in terms of ACC@5) for all the datasets for the three overlap metrics (JS, LCST, OFE)  for trajectories with a 0-20\%. We provide the results for all datasets and the other overlaps in Supplementary \ref{app:rerank}.}
    \label{fig:performances_scaled}
    
\end{figure*}

Table \ref{tab:augmented} and Figure \ref{fig:performances_scaled} show how the accuracy changes on the test trajectories with 0-20 overlap on all the datasets and models considered. 
Our reranking leads to improved accuracy regardless of the dataset and the overlap measures used. 
The bigger relative improvements are related to the trajectories with an overlap of 0-20. 
Regarding check-in datasets, on Foursquare New York, the improvement varies from +3.25\% (ST-RNN) to +9.38 (LSTPM). 
Similarly, on Foursquare Tokyo, the improvement varies from +5.69\% (DeepMove) to a +9.33\% of improvement (STAN). 
In Gowalla, we have the lowest relative improvement on DeepMove (+4.43\%) and the highest on RNN (+29.09\%).
Concerning taxi datasets, on Taxi Porto, the relative improvement on the average case (i.e., without stratifying the test set) varies from a +2.68\% (RNN) to +5.84\% (STAN). 
On Taxi San Francisco, the relative improvement varies from +2.49\% (RNN) to +5.74\% (DeepMove).
Regarding the 0-20 stratification, the largest relative improvement is associated with metrics JS, followed by LCST and OFE. 
On Foursquare New York, the relative improvement with JS is up to +96.15\%, with LCST being +20.39\%, and with OFE being +33.05\%. Similarly, on Foursquare Tokyo, we have top relative improvements of +82.35\%, +21.78\%, and +24.36\% with JS, LCTS, and OFE, respectively. Finally, Gowalla's top relative improvements for JS, LCTS, and OFE are +68.82\%, +45.45\%, and +50.03\%. 
In general, taxi datasets are associated with the lowest relative improvement: 
with JS, it is up to +7.96\%, with LCST +6.68\%, and with OFE +7.05\% on Taxi Porto. On the other hand, on Taxi San Francisco, the relative improvements for JS, LCST, and OFE are +5.82\%, +9.68\%, and +8.76\%.
The largest relative improvement is associated with the 0-20 overlap scenario. 
For example, the largest relative improvement on the 80-100 bin is 0.12\%. 
In other words, our rerank strategy brings the largest improvement in accuracy, especially where NLs are the least accurate.

\begin{table*}[]
\centering
\resizebox{\textwidth}{!}{%
\begin{tabular}{l|l|llll}
                                 & NL + RR                     &                                       & \textbf{JS}                                     & LCST                                   & OFE                                    \\ \hline
                                 & { RNN}  & { .233 (+9.38\%)} & { \textbf{.051 (+96.15\%)}} & { .158 (+19.69\%)} & { .241 (+26.17\%)} \\
                                 & ST-RNN                      & .261 (+5,24\%)                        & \textbf{.059 (+84.37\%)}                        & .186 (+15.52\%)                        & .299 (+31.14\%)                        \\
                                 & Deep Move                   & .277 (+6,94\%)                        & \textbf{.084 (+64.71\%)}                        & .213 (+19.66)                          & .268 (+16.01\%)                        \\
                                 & LSTPM                       & .272 (+8.36\%)                        & \textbf{.072 (+56.52\%)}                        & .184 (+10.84\%)                        & .271 (+18.34\%)                        \\
\multirow{-5}{*}{Foursquare NYC} & STAN                        & .281 (+6.43\%)                        & \textbf{.101 (+32.89\%)}                        & .214 (+15.05\%)                        & .283 (+18.90\%)                        \\ \hline

                                 & { RNN}  & { .196 (+6.56\%)} & { \textbf{.028 (+40.02\%)}} & { .123 (+21.78\%)} & { .171 (+15.54\%)} \\
                                 & ST-RNN                      & .213 (+7,58\%)                        & \textbf{.057 (+67.65\%)}                        & .133 (+15.64\%)                        & .194 (+24.36\%)                        \\
                                 & Deep Move                   & .223 (+5,69\%)                        & \textbf{.060 (+46.34\%)}                        & .142 (+19.33\%)                          & .201 (+19.64\%)                        \\
                                 & LSTPM                       & .233 (+6.88\%)                        & \textbf{.074 (+57.45\%)}                        & .151 (+21.77\%)                        & .236 (+34.86\%)                        \\
\multirow{-5}{*}{Foursquare TKY} & STAN                        & .246 (+9.33\%)                        & \textbf{.093 (+82.35\%)}                        & .153 (+18.60\%)                        & .239 (+32.04\%)                        \\ \hline

                                 & { RNN}  & { .142 (+29.09\%)} & { \textbf{.157 (+68.82\%)}} & { .016 (+45.45\%)} & { .144 (+50.03\%)} \\
                                 & ST-RNN                      & .149 (+8.76\%)                        & \textbf{.143 (+28.83\%)}                        & .033 (+17.86\%)                        & .127 (+19.81\%)                        \\
                                 & Deep Move                   & .165 (+4.43\%)                        & \textbf{.164 (+41.38\%)}                        & .041 (+13.89)                          & .151 (+32.46\%)                        \\
                                 & LSTPM                       & .171 (+12.50\%)                        & \textbf{.182 (+61.06\%)}                        & .043 (+34.38\%)                        & .151 (+36.04\%)                        \\
\multirow{-5}{*}{Gowalla} & STAN                        & .206 (+6.77\%)                        & \textbf{.178 (+43.55\%)}                        & .059 (+15.69\%)                        & .146 (+22.69\%)                        \\ \hline

                                 & RNN                         & .421 (+2.68\%)                        & \textbf{.069 (+4.54\%)}                         & .296 (+1.02\%)                         & .398 (+1.79\%)                         \\
                                 & ST-RNN                      & .427 (+2.64\%)                        & {.077 (+5.47\%)}                         & .313 (+3.98\%)                         & \textbf{.418 (+5.55\%)}                         \\
                                 & DeepMove                    & .466 (+5.42\%)                        & \textbf{.104 (+6.12\%)}                         & .341 (+3.96\%)                         & .434 (+6.11\%)                         \\
                                 & LSTPM                       & .457 (+6.52\%)                        & \textbf{.095 (+6.74\%)}                         & .336 (+6.32\%)                         & .419 (+5.01\%)                         \\
\multirow{-5}{*}{Taxi Porto}     & { STAN} & { .483 (+6.62\%)} & { \textbf{.111 (+7.96\%)}}  & { .351 (+6.68\%)}  & { .440 (+7.05\%)}  \\ \hline

                                 & { RNN}  & { .288 (+2.49\%)} & { {.193 (+4.89\%)}} & { .208 (+1.46\%)} & \textbf{{ .276 (+5.34\%)}} \\
                                 & ST-RNN                      & .297 (+4,95\%)                        & {.200 (+5.82\%)}                        & .225 (+6.64\%)                        & \textbf{.298 (+8.76\%) }                       \\
                                 & Deep Move                   & .313 (+5,74\%)                        & {.202 (+4.12\%)}                        & .227 (+4.13\%)                          & \textbf{.297 (+6.45\%)}                        \\
                                 & LSTPM                       & .301 (+5.24\%)                        & {.199 (+3.65\%)}                        & .238 (+9.68\%)                        & \textbf{.293 (+5.02\%)}                        \\
\multirow{-5}{*}{Taxi SF} & STAN                        & .330 (+5.11\%)                        & {.208 (+3.48\%)}                        & .233 (+4.48\%)                        & \textbf{.309 (+5.79\%)  }                      \\ \hline

\end{tabular}}
\caption{ACC@5 of all the models on all the datasets. 
We find a significant relative improvement, especially on the trajectories with a 0-20 overlap. 
Regarding check-in datasets, we have the greatest relative improvement on the stratification based on JS (in bold). In taxi datasets, we have similar improvements on JS and OFE and while in Taxi Porto, we have the best improvements on JS, on Taxi San Francisco, we reach the best improvements on OFE. In general, the improvements in check-in datasets are higher with respect to taxi datasets. There is not a specific model on which we have the best improvements.}
\label{tab:augmented}
\end{table*}

%{\color{red} We also use Shap \cite{shap} to understand better how the mobility laws impact the reranking procedure. The old score always plays the most important role regardless of the dataset. On the datasets related to taxis, the distance law is the most important when reranking, while on the check-in datasets, the distance law and the visitation law play similar roles. In both cases, the returner explorer law is the less considered.} 

\section{Discussion and Recommendations}

This work finds that the models' performances are deeply affected by the level of overlap present in the test trajectories. 
Based on the amount of trajectory overlap, we identify three scenarios: 

\begin{itemize}
    \item \textbf{Known Mobility}: the NL sees the entire trajectory in the training phase (overlap between 80\% and 100\%). 
    Predictive performance is much higher than the performance on a non-stratified test set (close to 100\%) as the test trajectories are almost identical to the training trajectories.
    
    \item \textbf{Fragmentary Mobility}: the NL sees a significant portion of the trajectory (overlap between 20\% and 80\%). 
    The majority of trajectories in the test set lies in this scenario. 
    There is a drop in the model performance compared to the previous scenario, decreasing up to $\sim$80\%.
    
    \item \textbf{Novel Mobility}: the NL sees a tiny or no portion of the trajectory (overlap below 20\%). 
    A significant number of trajectories lie in this scenario. However, since NLs cannot rely on the trajectories already seen in the training phase, these are the most difficult trajectories to predict. 
    Indeed, the performance of NLs on test sets with low overlap is considerably lower than the performance on a non-stratified test set.

\end{itemize}

While predicting known mobility is a simple task, inferring mobility patterns for fragmentary mobility and novel mobility presents challenges (e.g., dealing with under-represented locations or not represented at all in the training set). From a modeling perspective, this may suggest that current models are excellent in memorizing already seen trajectories but cannot generalize well. Some works suggest that reranking techniques or few-short learning algorithms may help solve this problem \cite{wang2020generalizing}. Also, results indicated that NLs might not be evaluated adequately.
In this sense, here we provide a set of recommendations for the evaluation of NLs:

\begin{enumerate}
    \item MMCs achieve performance similar to NLs. 
    Therefore, we claim that MMCs and other Markov chains approaches should always be used as a baseline.
    
    \item Although NLs achieve good overall performance, they are significantly biased due to trajectory overlap. 
    Besides the NLs' average performance, researchers should report the performance for the known mobility and novel mobility scenarios. 
    It is indeed crucial to understand whether the improved performance of the proposed NL is actually due to its increasing generalization capability or because it is memorizing better the trajectories in the training set;
    
    \item {\color{black} 
    NLs achieve the worst performance on the 0-20 overlap bin. 
    We can improve the performance on this bin, hence increasing NLs' generalization capability with the support of well-known spatial mobility laws, which are loosely captured by state-of-the-art NLs given their reliance on RNNs.}
\end{enumerate}
From other perspectives (e.g., urban planning, sustainability, and others), having models that can generalize well is fundamentally important. First, NLs that generalize can be used to perform better simulations and to analyze what-if scenarios more realistically. For instance, we may be able to see how attractive a new POI in a specific place would be. We cannot solve such problems with a model that only memorizes seen trajectories. Also, it can help urban planners make decisions about traffic and transportation and, thus, reduce pollution. We can also use an NL that can generalize to predict better and understand the mobility of individuals who have never been seen in a region (e.g., a tourist). Also, a generalized model may be geographically transferable (e.g., trained in an area and tested on a new territory). This may represent a significant step toward solutions to some of the United Nations' Sustainable Development Goals. In particular, we may use such models to run simulations or investigate pollution, inclusion, and the design of better cities in territories where we do not have data or have a scarcity of data. 

\section{Conclusions}
In this work, we investigate the generalization capabilities of next-location prediction datasets. 
We find that model performance is considerably affected by trajectory test-train overlap, suggesting that NLs memorize training trajectories rather than generalizing.
We show we mitigate this issue by injecting mobility laws into state-of-the-art NLs, achieving relative improvement on test sets with low overlap with the training ones.
We aim to consider other mobility laws and use more sophisticated models to rerank the results in future work.
It would also be helpful to use explainable AI techniques to understand better the role of mobility laws and the relations between the DL modules composing NLs.

\section*{Declarations}

\noindent \textbf{Funding}
Luca Pappalardo has been partially supported by EU project SoBigData++ grant agreement 871042.

\noindent \textbf{Conflict of Interest}
The authors have no competing interests to declare that are relevant to the content of this article.

\noindent \textbf{Ethics approval} not applicable

\noindent \textbf{Consent to participate} not applicable

\noindent \textbf{Consent for publication} not applicable

\noindent \textbf{Availability of data and materials}
All the data are publicly available and can be downloaded using the links at \href{github.com/scikit-mobility/DeepLearning4HumanMobility}{https://github.com/scikit-mobility/DeepLearning4HumanMobility}.

\noindent \textbf{Code availability} The code used to compute the overlap can be found at \href{github.com/MassimilianoLuca/overlap-processing}{https://github.com/MassimilianoLuca/overlap-processing} the code of the models can be found at \href{github.com/LibCity/Bigscity-LibCity}{https://github.com/LibCity/Bigscity-LibCity}

\noindent \textbf{Authors' contributions}
M.L. designed the methodology to compute the overlap and the rerank methodology. G.B. directed the study. All the authors contributed to interpreting the results and writing the paper. G.B. developed this work prior joining Amazon.


\newpage
\begin{appendices}

\section{NL's performances}
\label{app:perf}

\begin{table}[htb]
\centering
\resizebox{\textwidth}{!}{%
\begin{tabular}{lll|lllll|lllll|lllll|}
                                &                               & \multicolumn{1}{c|}{} & \multicolumn{5}{c|}{JC}                & \multicolumn{5}{c|}{LCST}               & \multicolumn{5}{c|}{OFE}               \\
                                &                               &                       & 0-20  & 20-40 & 40-60 & 60-80 & 80-100 & 0-20  & 20-40 & 40-60 & 60-80 & 80-100  & 0-20  & 20-40 & 40-60 & 60-80 & 80-100 \\ \hline
\multirow{6}{*}{Gowalla}        & \multicolumn{1}{l|}{MMC}      & 0.129                 & 0.108 & 0.119 & 0.323 & 0.526 & 0.776  & 0.013 & 0.182 & 0.398 & 0.644 & 0.837   & 0.099 & 0.211 & 0.546 & 0.783 & 0.874  \\
                                & \multicolumn{1}{l|}{RNN}      & 0.110                 & 0.093 & 0.109 & 0.321 & 0.498 & 0.764  & 0.011 & 0.173 & 0.385 & 0.631 & 0.824   & 0.096 & 0.209 & 0.532 & 0.777 & 0.871  \\
                                & \multicolumn{1}{l|}{ST-RNN}   & 0.137                 & 0.111 & 0.128 & 0.355 & 0.513 & 0.782  & 0.028 & 0.195 & 0.406 & 0.658 & 0.849   & 0.106 & 0.217 & 0.558 & 0.793 & 0.888  \\
                                & \multicolumn{1}{l|}{DeepMove} & 0.158                 & 0.116 & 0.142 & 0.373 & 0.536 & 0.798  & 0.036 & 0.211 & 0.434 & 0.681 & 0.869   & 0.114 & 0.226 & 0.573 & 0.822 & 0.899  \\
                                & \multicolumn{1}{l|}{LSTPM}    & 0.152                 & 0.113 & 0.141 & 0.369 & 0.528 & 0.793  & 0.032 & 0.203 & 0.427 & 0.669 & 0.861   & 0.111 & 0.219 & 0.569 & 0.818 & 0.897  \\
                                & \multicolumn{1}{l|}{STAN}     & 0.192                 & 0.124 & 0.183 & 0.398 & 0.561 & 0.813  & 0.051 & 0.221 & 0.449 & 0.704 & 0.888   & 0.119 & 0.231 & 0.579 & 0.829 & 0.911  \\ \hline
\multirow{6}{*}{Foursquare NYC} & \multicolumn{1}{l|}{MMC}      & 0.245                 & 0.045 & 0.214 & 0.518 & 0.730 & 0.889  & 0.102 & 0.336 & 0.653 & 0.858 & 0.987   & 0.230 & 0.529 & 0.778 & 0.885 & 0.988  \\
                                & \multicolumn{1}{l|}{RNN}      & 0.213                 & 0.026 & 0.186 & 0.472 & 0.643 & 0.740  & 0.132 & 0.271 & 0.504 & 0.528 & 0.869   & 0.191 & 0.610 & 0.712 & 0.926 & 0.988  \\
                                & \multicolumn{1}{l|}{ST-RNN}   & 0.248                 & 0.032 & 0.236 & 0.538 & 0.709 & 0.859  & 0.161 & 0.322 & 0.544 & 0.589 & 0.920   & 0.228 & 0.679 & 0.779 & 0.928 & 0.989  \\
                                & \multicolumn{1}{l|}{DeepMove} & 0.259                 & 0.051 & 0.241 & 0.539 & 0.716 & 0.873  & 0.178 & 0.330 & 0.569 & 0.631 & 0.926   & 0.231 & 0.683 & 0.786 & 0.935 & 0.996  \\
                                & \multicolumn{1}{l|}{LSTPM}    & 0.251                 & 0.046 & 0.239 & 0.533 & 0.717 & 0.865  & 0.166 & 0.327 & 0.551 & 0.599 & 0.922   & 0.229 & 0.680 & 0.782 & 0.929 & 0.991  \\
                                & \multicolumn{1}{l|}{STAN}     & 0.264                 & 0.076 & 0.269 & 0.544 & 0.732 & 0.875  & 0.186 & 0.331 & 0.574 & 0.636 & 0.925 & 0.238 & 0.686 & 0.792 & 0.936 & 0.992  \\ \hline
\multirow{6}{*}{Foursquare TKY} & \multicolumn{1}{l|}{MMC}      & 0.216                 & 0.043 & 0.172 & 0.450 & 0.691 & 0.856  & 0.096 & 0.243 & 0.428 & 0.634 & 0.718   & 0.199 & 0.360 & 0.691 & 0.716 & 0.991  \\
                                & \multicolumn{1}{l|}{RNN}      & 0.183                 & 0.020 & 0.153 & 0.390 & 0.609 & 0.783  & 0.101 & 0.205 & 0.381 & 0.593 & 0.692   & 0.148 & 0.464 & 0.703 & 0.737 & 0.892  \\
                                & \multicolumn{1}{l|}{ST-RNN}   & 0.198                 & 0.034 & 0.158 & 0.397 & 0.637 & 0.794  & 0.115 & 0.224 & 0.394 & 0.608 & 0.699   & 0.156 & 0.468 & 0.711 & 0.739 & 0.906  \\
                                & \multicolumn{1}{l|}{LSTPM}    & 0.211                 & 0.041 & 0.164 & 0.431 & 0.688 & 0.848  & 0.119 & 0.233 & 0.419 & 0.621 & 0.715   & 0.168 & 0.471 & 0.718 & 0.743 & 0.931  \\
                                & \multicolumn{1}{l|}{DeepMove} & 0.218                 & 0.047 & 0.168 & 0.452 & 0.694 & 0.854  & 0.124 & 0.239 & 0.424 & 0.638 & 0.719   & 0.175 & 0.472 & 0.723 & 0.744 & 0.935  \\
                                & \multicolumn{1}{l|}{STAN}     & 0.225                 & 0.051 & 0.174 & 0.458 & 0.696 & 0.857  & 0.129 & 0.244 & 0.426 & 0.639 & 0.722   & 0.181 & 0.475 & 0.728 & 0.749 & 0.938  \\ \hline
\multirow{6}{*}{Taxi Porto}     & \multicolumn{1}{l|}{MMC}      & 0.309                 & 0.113 & 0.258 & 0.316 & 0.385 & 0.482  & 0.274 & 0.302 & 0.343 & 0.397 & 0.411   & 0.286 & 0.333 & 0.349 & 0.427 & 0.436  \\
                                & \multicolumn{1}{l|}{RNN}      & 0.410                 & 0.066 & 0.269 & 0.398 & 0.516 & 0.638  & 0.293 & 0.381 & 0.461 & 0.542 & 0.649   & 0.391 & 0.458 & 0.516 & 0.617 & 0.643  \\
                                & \multicolumn{1}{l|}{ST-RNN}   & 0.416                 & 0.073 & 0.272 & 0.404 & 0.519 & 0.639  & 0.301 & 0.395 & 0.467 & 0.544 & 0.653   & 0.396 & 0.463 & 0.518 & 0.623 & 0.649  \\
                                & \multicolumn{1}{l|}{DeepMove} & 0.442                 & 0.098 & 0.296 & 0.419 & 0.528 & 0.651  & 0.328 & 0.417 & 0.486 & 0.561 & 0.672   & 0.409 & 0.467 & 0.532 & 0.637 & 0.684  \\
                                & \multicolumn{1}{l|}{LSTPM}    & 0.429                 & 0.089 & 0.281 & 0.413 & 0.521 & 0.647  & 0.316 & 0.411 & 0.473 & 0.558 & 0.668   & 0.399 & 0.466 & 0.524 & 0.629 & 0.661  \\
                                & \multicolumn{1}{l|}{STAN}     & 0.453                 & 0.103 & 0.305 & 0.421 & 0.532 & 0.649  & 0.329 & 0.423 & 0.492 & 0.558 & 0.679   & 0.411 & 0.474 & 0.537 & 0.631 & 0.692  \\ \hline
\multirow{6}{*}{Taxi SF}        & \multicolumn{1}{l|}{MMC}      & 0.242                 & 0.169 & 0.261 & 0.385 & 0.542 & 0.583  & 0.194 & 0.248 & 0.308 & 0.451 & 0.505   & 0.231 & 0.256 & 0.389 & 0.427 & 0.692  \\
                                & \multicolumn{1}{l|}{RNN}      & 0.281                 & 0.184 & 0.279 & 0.403 & 0.565 & 0.601  & 0.205 & 0.263 & 0.321 & 0.475 & 0.539   & 0.262 & 0.303 & 0.414 & 0.504 & 0.816  \\
                                & \multicolumn{1}{l|}{ST-RNN}   & 0.283                 & 0.189 & 0.284 & 0.411 & 0.568 & 0.607  & 0.211 & 0.274 & 0.329 & 0.486 & 0.549   & 0.274 & 0.315 & 0.429 & 0.521 & 0.834  \\
                                & \multicolumn{1}{l|}{LSTPM}    & 0.296                 & 0.194 & 0.286 & 0.429 & 0.574 & 0.609  & 0.218 & 0.278 & 0.334 & 0.492 & 0.558   & 0.279 & 0.319 & 0.438 & 0.529 & 0.847  \\
                                & \multicolumn{1}{l|}{DeepMove} & 0.286                 & 0.192 & 0.287 & 0.414 & 0.571 & 0.609  & 0.217 & 0.277 & 0.332 & 0.489 & 0.551   & 0.279 & 0.318 & 0.435 & 0.527 & 0.841  \\
                                & \multicolumn{1}{l|}{STAN}     & 0.314                 & 0.201 & 0.295 & 0.433 & 0.581 & 0.614  & 0.223 & 0.281 & 0.349 & 0.498 & 0.573   & 0.293 & 0.326 & 0.443 & 0.532 & 0.850  \\ \hline
\end{tabular}%
}
\caption{ACC@5 of all the models on all the datasets without a stratification (first column) and with the train-test stratification based on overlap metric and percentage of overlap}
\label{tab:acc_5_complete}
\end{table}

\section{NL's performances after re-ranking}
\label{app:rerank}
\begin{table*}[htb]
\centering
\resizebox{\textwidth}{!}{%
\begin{tabular}{lll|lllll|lllll|lllll|}
\multicolumn{1}{c}{}        & \multicolumn{1}{c}{}          & \multicolumn{1}{c|}{} & \multicolumn{5}{c|}{JC}                & \multicolumn{5}{c|}{LCST}               & \multicolumn{5}{c|}{OFE}               \\
                            &                               &                       & 0-20  & 20-40 & 40-60 & 60-80 & 80-100 & 0-20  & 20-40 & 40-60 & 60-80 & 80-100  & 0-20  & 20-40 & 40-60 & 60-80 & 80-100 \\ \hline
\multirow{5}{*}{Gowalla}    & \multicolumn{1}{l|}{RNN}      & 0.140                 & 0.160 & 0.148 & 0.347 & 0.533 & 0.777  & 0.016 & 0.192 & 0.403 & 0.647 & 0.837   & 0.144 & 0.257 & 0.546 & 0.783 & 0.874  \\
                            & \multicolumn{1}{l|}{ST-RNN}   & 0.117                 & 0.143 & 0.148 & 0.326 & 0.506 & 0.765  & 0.013 & 0.184 & 0.393 & 0.634 & 0.825   & 0.127 & 0.263 & 0.532 & 0.777 & 0.871  \\
                            & \multicolumn{1}{l|}{DeepMove} & 0.145                 & 0.164 & 0.156 & 0.371 & 0.521 & 0.783  & 0.033 & 0.204 & 0.415 & 0.661 & 0.850   & 0.151 & 0.265 & 0.558 & 0.793 & 0.888  \\
                            & \multicolumn{1}{l|}{LSTPM}    & 0.171                 & 0.182 & 0.178 & 0.379 & 0.544 & 0.799  & 0.043 & 0.225 & 0.452 & 0.684 & 0.869   & 0.151 & 0.276 & 0.573 & 0.822 & 0.899  \\
                            & \multicolumn{1}{l|}{STAN}     & 0.163                 & 0.178 & 0.182 & 0.385 & 0.615 & 0.797  & 0.039 & 0.209 & 0.440 & 0.670 & 0.862   & 0.146 & 0.267 & 0.569 & 0.818 & 0.897  \\ \hline
\multirow{5}{*}{Foursquare NYC}      & \multicolumn{1}{l|}{RNN}      & 0.233                 & 0.051 & 0.265 & 0.479 & 0.651 & 0.741  & 0.158 & 0.276 & 0.598 & 0.531 & 0.869   & 0.241 & 0.742 & 0.712 & 0.926 & 0.988  \\
                            & \multicolumn{1}{l|}{ST-RNN}   & 0.261                 & 0.059 & 0.287 & 0.563 & 0.722 & 0.861  & 0.186 & 0.328 & 0.552 & 0.590 & 0.920   & 0.299 & 0.787 & 0.779 & 0.928 & 0.989  \\
                            & \multicolumn{1}{l|}{DeepMove} & 0.277                 & 0.084 & 0.300 & 0.571 & 0.724 & 0.875  & 0.213 & 0.358 & 0.593 & 0.633 & 0.927   & 0.268 & 0.766 & 0.786 & 0.935 & 0.996  \\
                            & \multicolumn{1}{l|}{LSTPM}    & 0.272                 & 0.072 & 0.290 & 0.562 & 0.728 & 0.873  & 0.184 & 0.356 & 0.576 & 0.601 & 0.923   & 0.271 & 0.763 & 0.782 & 0.929 & 0.991  \\
                            & \multicolumn{1}{l|}{STAN}     & 0.281                 & 0.101 & 0.327 & 0.570 & 0.740 & 0.877  & 0.214 & 0.359 & 0.592 & 0.638 & 0.925 & 0.283 & 0.778 & 0.792 & 0.936 & 0.992  \\ \hline
\multirow{5}{*}{Foursquare TKY}      & \multicolumn{1}{l|}{RNN}      & 0.195                 & 0.028 & 0.186 & 0.408 & 0.618 & 0.785  & 0.123 & 0.317 & 0.394 & 0.596 & 0.692   & 0.171 & 0.520 & 0.703 & 0.737 & 0.892  \\
                            & \multicolumn{1}{l|}{ST-RNN}   & 0.213                 & 0.057 & 0.213 & 0.411 & 0.647 & 0.800  & 0.133 & 0.235 & 0.408 & 0.614 & 0.699   & 0.194 & 0.525 & 0.711 & 0.739 & 0.906  \\
                            & \multicolumn{1}{l|}{LSTPM}    & 0.223                 & 0.060 & 0.208 & 0.452 & 0.698 & 0.856  & 0.142 & 0.246 & 0.437 & 0.622 & 0.715   & 0.200 & 0.540 & 0.718 & 0.743 & 0.931  \\
                            & \multicolumn{1}{l|}{DeepMove} & 0.230                 & 0.074 & 0.209 & 0.488 & 0.704 & 0.858  & 0.151 & 0.338 & 0.444 & 0.643 & 0.719   & 0.236 & 0.564 & 0.723 & 0.744 & 0.935  \\
                            & \multicolumn{1}{l|}{STAN}     & 0.236                 & 0.093 & 0.278 & 0.501 & 0.707 & 0.865  & 0.153 & 0.329 & 0.447 & 0.642 & 0.726   & 0.239 & 0.553 & 0.728 & 0.749 & 0.938  \\ \hline
\multirow{5}{*}{Taxi Porto} & \multicolumn{1}{l|}{RNN}      & 0.421                 & 0.069 & 0.275 & 0.436 & 0.525 & 0.643  & 0.306 & 0.387 & 0.468 & 0.545 & 0.649   & 0.398 & 0.458 & 0.516 & 0.617 & 0.643  \\
                            & \multicolumn{1}{l|}{ST-RNN}   & 0.427                 & 0.077 & 0.279 & 0.415 & 0.529 & 0.641  & 0.317 & 0.401 & 0.473 & 0.548 & 0.654   & 0.418 & 0.463 & 0.518 & 0.623 & 0.649  \\
                            & \multicolumn{1}{l|}{LSTPM}    & 0.466                 & 0.104 & 0.301 & 0.429 & 0.536 & 0.652  & 0.348 & 0.426 & 0.498 & 0.563 & 0.673   & 0.434 & 0.467 & 0.532 & 0.637 & 0.684  \\
                            & \multicolumn{1}{l|}{DeepMove} & 0.457                 & 0.095 & 0.287 & 0.423 & 0.529 & 0.650  & 0.337 & 0.422 & 0.482 & 0.559 & 0.668   & 0.419 & 0.466 & 0.524 & 0.629 & 0.661  \\
                            & \multicolumn{1}{l|}{STAN}     & 0.483                 & 0.111 & 0.316 & 0.436 & 0.542 & 0.652  & 0.355 & 0.431 & 0.502 & 0.563 & 0.680   & 0.440 & 0.474 & 0.537 & 0.631 & 0.692  \\ \hline
\multirow{5}{*}{Taxi SF}    & \multicolumn{1}{l|}{RNN}      & 0.288                 & 0.193 & 0.285 & 0.422 & 0.571 & 0.604  & 0.208 & 0.271 & 0.327 & 0.477 & 0.539   & 0.276 & 0.303 & 0.414 & 0.504 & 0.816  \\
                            & \multicolumn{1}{l|}{ST-RNN}   & 0.297                 & 0.200 & 0.294 & 0.433 & 0.575 & 0.608  & 0.225 & 0.278 & 0.333 & 0.488 & 0.549   & 0.298 & 0.315 & 0.429 & 0.521 & 0.834  \\
                            & \multicolumn{1}{l|}{DeepMove} & 0.313                 & 0.202 & 0.292 & 0.449 & 0.581 & 0.610  & 0.230 & 0.282 & 0.338 & 0.493 & 0.558   & 0.297 & 0.319 & 0.438 & 0.529 & 0.847  \\
                            & \multicolumn{1}{l|}{LSTPM}    & 0.301                 & 0.199 & 0.297 & 0.433 & 0.578 & 0.613  & 0.238 & 0.282 & 0.338 & 0.493 & 0.552   & 0.293 & 0.318 & 0.435 & 0.527 & 0.841  \\
                            & \multicolumn{1}{l|}{STAN}     & 0.330                 & 0.208 & 0.304 & 0.461 & 0.592 & 0.617  & 0.233 & 0.288 & 0.355 & 0.503 & 0.574   & 0.310 & 0.326 & 0.443 & 0.532 & 0.850  \\ \hline
\end{tabular}%
}
\caption{ACC@5 of all the models after the re-ranking on all the datasets without a stratification (first column) and with the train-test stratification
based on overlap metric and percentage of overlap
}
\label{tab:my-table}
\end{table*}

\end{appendices}

%\bibliographystyle{plain}
\bibliography{sn-bibliography}% common bib file


\end{document}